\documentclass[runningheads]{llncs}

\usepackage[colorlinks]{hyperref}

\usepackage{amsmath}
\usepackage{amssymb}
\usepackage{CJK}
\usepackage{url}
\usepackage{float}
\usepackage{graphicx}
\usepackage{diagbox}
\usepackage{multirow}
\usepackage{wrapfig}
\usepackage[]{graphbox}

\usepackage{algorithm}
\usepackage{algpseudocode}
\usepackage{booktabs}
\usepackage{subfigure}
\usepackage{tikz} 
\usetikzlibrary{shadows,arrows,shapes}
\pgfdeclarelayer{background}
\pgfdeclarelayer{foreground}
\pgfsetlayers{background,main,foreground}
 
\begin{document}


\title {Multi-Stage Prediction Networks for\\ Data Harmonization}
\titlerunning{Multi-Stage Prediction Networks for Data Harmonization}
\author{Stefano B. Blumberg\inst{1} \and Marco Palombo\inst{1} \and Can Son Khoo\inst{1} \and \\ Chantal M. W. Tax\inst{2} \and Ryutaro Tanno\inst{1} \and Daniel C. Alexander\inst{1}}

\authorrunning{Stefano B. Blumberg et al.}

\institute{Department of Computer Science and Centre for Medical Image Computing, \\ University College London (UCL)
\and Cardiff University Brain Research Imaging Centre (CUBRIC), Cardiff University
\\ \email{stefano.blumberg.17@ucl.ac.uk}}
\maketitle    

\begin{abstract}
In this paper, we introduce multi-task learning (MTL) to data harmonization (DH); where we aim to harmonize images across different acquisition platforms and sites.  This allows us to integrate information from multiple acquisitions and improve the predictive performance and learning efficiency of the harmonization model. Specifically, we introduce the Multi Stage Prediction (MSP) Network, a MTL framework that incorporates neural networks of potentially disparate architectures, trained for different individual acquisition platforms, into a larger architecture that is refined in unison.  The MSP utilizes high-level features of single networks for individual tasks, as inputs of additional neural networks to inform the final prediction, therefore exploiting redundancy across tasks to make the most of limited training data.  We validate our methods on a dMRI harmonization challenge dataset,  where we predict three modern platform types, from one obtained from an old scanner.  We show how MTL architectures, such as the MSP, produce around 20\% improvement of patch-based mean-squared error over current state-of-the-art methods and that our MSP outperforms off-the-shelf MTL networks.  
Our code is available \cite{CODE}.

\begin{keywords}
Data Harmonization, Deep Learning, Diffusion Magnetic Resonance Imaging, Multi-Task Learning, Transfer Learning.
\end{keywords}

\end{abstract}

\section{Introduction}\label{INTRO}

Lack of standardization amongst imaging acquisitions is a long-standing problem, that confounds large scale multi-centre imaging studies.  DH aims to remove differences arising from specifics of scanner, centre or acquisition protocol and increase the power of group studies.  The problem aligns with the wider issue of estimating, from an image obtained from a particular subject and scanner, the image that would have been obtained from the same subject on another scanner.  Various recent studies, such as Image Quality Transfer \cite{IQTVRFR,BIQT,DIQT} and Modality Transfer \cite{MPCSSSS}, present solutions to this problem that may be repurposed for DH.
\\ \indent Current approaches used in practice, such as \cite{HCTM,MSHDMRI}, generally register all images to a common template and align the mean and variances of image intensities from each platform (i.e. centre, scanner, and/or acquisition protocol), either voxelwise, regionally, or of whole images.     Recently, deep learning has shown much promise, e.g. in the Multi-Shell Data Harmonization Challenge (MUSHAC) \cite{CSCPDMRIDH,CSCPHMSDMRIOCEV,MSDMRIEC}, outperforming statistical approaches. Methods to date include the Convolutional Neural Network with Rotationally Invariant Spherical Harmonics (CNN-RISH) \cite{HDMRID} -- a 5-layer CNN that harmonizes dMRI from 3T to 7T; the Deeper Image Quality Transfer Network (DIQT) \cite{DIQT} -- an extension of the FCSNet \cite{CSCPDMRIDH}, which holds state-of-the-art results in dMRI super-resolution and the Spherical Harmonic Residual Network (SHResNet) \cite{SHRN} -- A 7-layer CNN that processes different spherical harmonic (SH) coefficients separately.  A common feature of all these methods is that they use a separate single CNN to estimate the target images from those of each individual platforms. 
\\ \indent Multi-scanner or multi-centre studies often acquire traveling heads (TH) -- a small number of subjects specifically scanned at each site, to support DH learning mappings from scanner to scanner. These learned mappings then transform other subjects to a common representation.  However, acquiring TH data is expensive and typically the number of subjects is small (rarely more than 10).  Therefore, single-network deep learning approaches easily overfit \cite{DIQT,SHRN} and fail to exploit the synergy between multiple, often strongly related, prediction tasks.    
\\ \indent MTL potentially offers a powerful solution to the paucity of training data in DH, by allowing the model to integrate information from acquisitions from multiple platforms. In contrast to single-network approaches, it exploits commonalities and differences across multiple tasks (in our case, predicting different acquisitions) to improve the predictive power of neural networks.  Furthermore by incorporating predictions and loss functions into a single network, MTL approaches acquire additional regularisation \cite{DSN}.  There are a variety of existing approaches that cascade over multiple predictions of varying resolutions, that might be adapted to a MTL approach in DH.  Examples include the Convolutional Pose Machine Networks (CPM) \cite{CPM} comprising a sequential architecture of CNNs, to provide increasing refined estimates and the Holistically-Nested Edge Detection Network (HNED) \cite{HNED} -- a single-stream deep network with multiple side outputs to perform multi-scale and multi-level feature learning.  However, these approaches process tasks sequentially, whilst the distinct tasks in DH differ in terms of resolution, contrast and noise patterns, with no simple notion of ordering.  Furthermore it is unclear as to what might be the optimal subnetworks for such a larger network.
\\ \indent  In this paper, we introduce a new MSP architecture designed to exploit, for the first time, MTL in DH.  The MSP draws on sub-network structures from \cite{CPM,HNED}, but cascades across various estimation tasks, rather than assuming they are sequentially structured.  This avoids both duplicating features and allows us to use redundancy across tasks to make the most of limited training data.
Furthermore the MSP allows us to incorporate pre-trained state-of-the art networks into a MTL framework, letting us explicitly use high-level features of multiple predictions, as network inputs to inform the final prediction.
\\ \indent  We utilize the recent MUSHAC \cite{CSCPDMRIDH} data to evaluate various strategies.  The specific task is to predict acquisitions of both low and high quality from different scanners, from data from the same subjects acquired on an ageing scanner.  We compare performance of the state of the art, i.e. various single network approaches, with MTL frameworks, including the MSP, where the best performing single networks inform construction of a particular MSP architecture.   The MSP outperforms both single networks and simple off-the-shelf MTL approaches.  We expect this initial demonstration to motivate wider development and usage of MTL strategies for DH, therefore we release our implementation in \cite{CODE}.

\section{Methods}\label{METHODS}

\begin{figure}[t]
\begin{center}
\includegraphics[width=\linewidth]{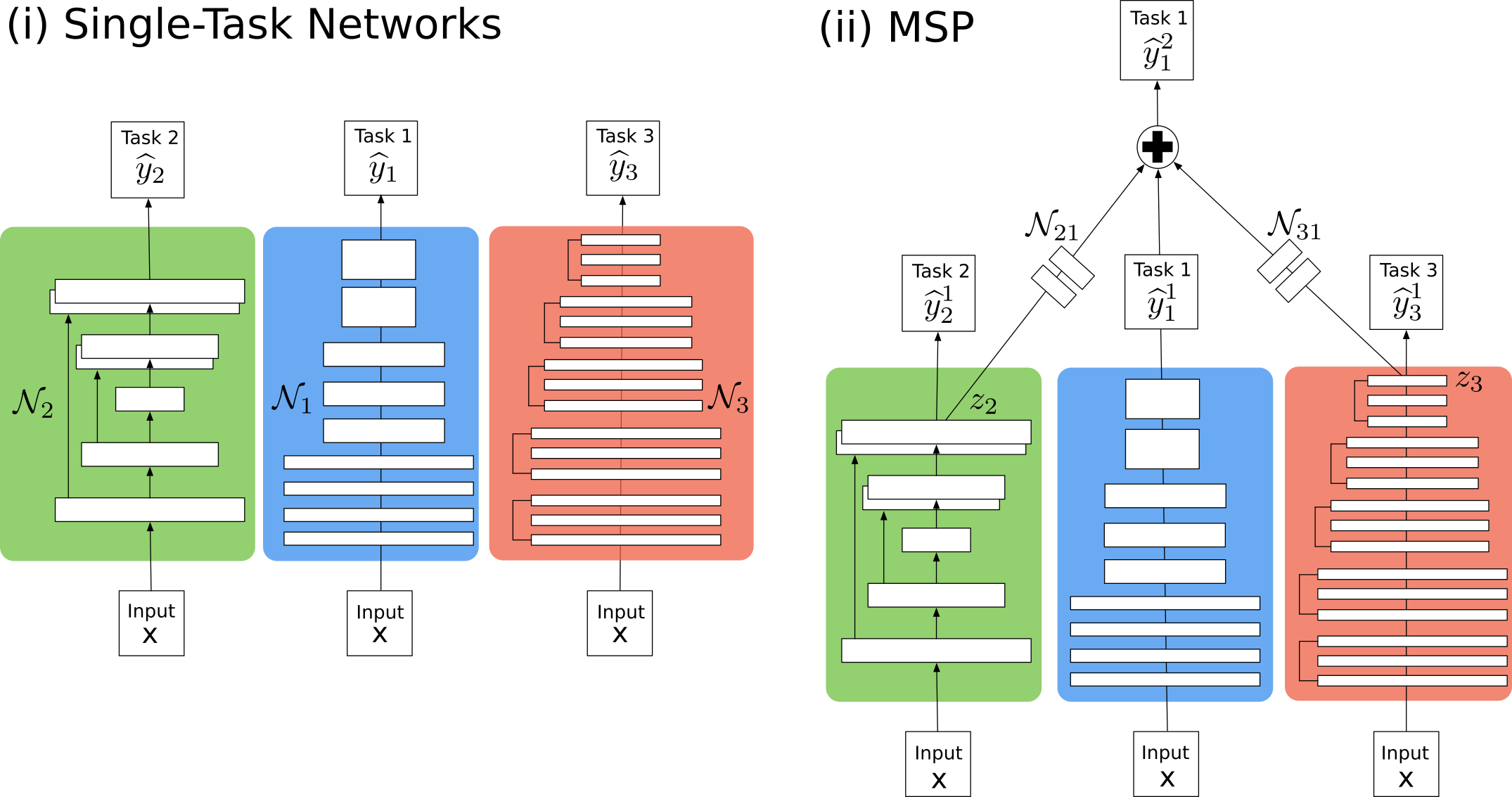}\caption{We illustrate creating the MSP from three single networks, with input, target platforms $ 0,1 $ and other platforms $ 2,3 $.  i) Three trained neural networks $ \mathcal{N}_{i}, \ i=1,2,3 $ separately predict patches $ \widehat{y}_{i} \ i=1,2,3 $, from input patch $ x $.  ii) The MSP.  We take $ \mathcal{N}_{2},\mathcal{N}_{3} $ and select their last features $ z_{i} \ i=2,3 $ as inputs to additional respective neural networks $ \mathcal{N}_{21},\mathcal{N}_{31} $.  The first-stage predictions are $ \widehat{y}^{1}_{i} \ i =1,2,3 $, the second-stage prediction,  a linear combination of $ \widehat{y}^{1}_{1}, \mathcal{N}_{21}(z_{2}),\mathcal{N}_{31}(z_{3}) $, is $ \widehat{y}^{2}_{1} $.}\label{SOA_vs_MSP}
\end{center}
\end{figure}

In this section we outline the general DH problem with THs and setting from MUSHAC \cite{CSCPDMRIDH} with which we validate our methods.  Then we construct the MSP.
\\
\\ In the typical DH setting, THs provide image data sets $ I_{ij} \ i=0,...,P-1 \ j=1,...,N $ for each of $ N $ subjects imaged on each of $ P $ platforms (a scanner and imaging protocol combination).  The challenge is to construct mappings $ M_{iT} \ i=0,...,P-1 $ from images acquired on platform $ i $ to the corresponding image acquired on some target platform $ T \in \{0,...,P-1\} $. 
\\ \indent MUSHAC \cite{CSCPDMRIDH,CSCPHMSDMRIOCEV,MSDMRIEC} tests the ability to predict multiple high quality targets (modern platforms) from a single low-quality input (an old out-of-date platform) thus concentrating on the most challenging aspect of DH.  With the same notation as before, we predict images $ I_{ij} \ i=1,...,P-1 \ j=1,...,N $ from $ I_{0j} \ j=1,..,N $. %
\\ \indent When predicting acquisitions from platform $ T \in \{1,...,P-1\} $, current state-of-the-art deep learning approaches extract patch pairs $ \{ (x(k),y_{T}(k)),... \} $ from respective $ I_{0j},I_{Tj} \ j=1,...,N $.  These approaches train a neural network $ \mathcal{N}_{T} $ on these patches, where for input patch $ x $ we denote the neural network's estimate (also denoted as prediction) of $ y_{T} $ as $ \widehat{y}_{T}$ i.e. $ \widehat{y}_{T} = \mathcal{N}_{T}(x) $ and the loss is calculated as $ L(\widehat{y}_{T},y_{T}) $.  This process is repeated for different $ T $.  In this paper we use CNN-RISH \cite{HDMRID}, DIQT \cite{DIQT}, SHResNet \cite{SHRN} to represent this class of technique.
\\ \indent In figure-\ref{SOA_vs_MSP} we contrast the MSP architecture that we propose, with the single network approach.  The MSP integrates information from multiple
platforms and utilizes high-level features from individual networks, to inform the final prediction with additional neural networks.  In training we use the set of corresponding patches
$ \{ (x(k),y_{1}(k),...,y_{T}(k),...),... \} $ from all images.  We first take the pre-trained networks $ \mathcal{N}_{i} \ i=1,...,P-1 $ as the state-of-the-art single-networks, which may have different architectures, that have already been optimized.  We denote for input patch $ x $, the predictions of these networks as the first-stage prediction of the MSP: $ \widehat{y}^{1}_{i} = \mathcal{N}_{i}(x) \ i = 1,...T $.  For networks $ \mathcal{N}_{i} \ i \neq T $, we denote its last feature map as $ z_{i} $.  We then create the new networks $ \mathcal{N}_{iT} \ i \neq T $, each taking input $ z_{i} $, where $ \mathcal{N}_{iT} $ informs the final prediction of platform $ T $ with information from platform $ i \neq T $.  The second-stage prediction of the MSP, is a linear combination of network outputs of $ \mathcal{N}_{iT}(z_{i}) \ i \neq T $, with the fist-stage prediction $ \widehat{y}^{1}_{T} $:
\begin{equation}
\widehat{y}^{2}_{T} =  (1-\alpha)\widehat{y}^{1}_{T} + \frac{\alpha}{P-1} (\sum_{i \neq T} \mathcal{N}_{iT}(z_{i}) + \widehat{y}^{1}_{T}) \,\ \ \alpha \in [0,1].    
\end{equation}
We train the whole MSP in unison, using supervised loss functions calculated at both stage predictions: $ L(\widehat{y}^{1}_{i},y_{i}),L(\widehat{y}^{2}_{T},y_{T}) \ i=1,...,P-1 $, which are backpropagated through the network.  For $ \alpha \in [0,1] $, we manually increase $ \alpha $ from $ 0 $ to $ 1 $ during training, to combine pre-trained $ \mathcal{N}_{T} $ with the untrained $ \mathcal{N}_{iT} $ as in \cite{PGG}.
\\ \indent  We provide an example of the MSP in figure-\ref{SOA_vs_MSP} for $ P=4, T=1 $ and code \cite{CODE}.

\section{Experiments and Results}\label{RESULTS}

\begin{figure}[t]
\begin{center}
\includegraphics[scale=0.5]{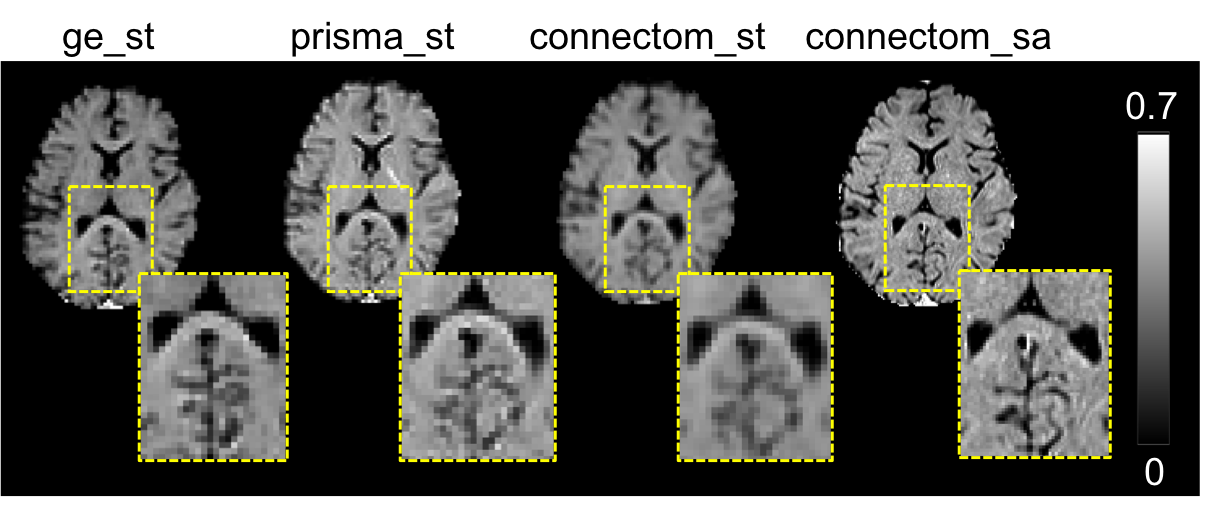}
\caption{Example of the normalized and direction-averaged dMRI image obtained from the same subject's brain using different acquisitions (st and sa) and different MRI scanners (GE, Prisma, Connectom). }\label{CDM}
\end{center}
\end{figure}

In this section we present an overview of our MUSHAC dMRI hamonization dataset \cite{CSCPDMRIDH}.  We describe how to process the raw data to produce patches for our supervised neural network approach.  We then illustrate how MTL approaches, such as the MSP, produce improved results over state-of-the-art methods.
\\
\\ \textbf{Harmonization Data}  Data are obtained from the MUSHAC \cite{CSCPDMRIDH}. 10 healthy volunteers were scanned on three different scanners, an ageing 3T General Electric Excite-HDx scanner (max. gradient 40 mT/m), a modern but standard 3T Siemens Prisma scanner (max. gradient 80 mT/m) and a state-of-the-art bespoke 3T Siemens Connectom scanner (max. gradient 300 mT/m). dMRI images were acquired for b = 1200 s/mm\textsuperscript{2} with two different manufacturer quality protocols: a standard (st) protocol -- with voxel side 2.4 mm, 30 directions per b-value, TE = 89 ms from all scanners; a state-of-the-art (sa) protocol on the Connectom -- with voxel size 1.2 mm, 60 directions per b-value, TE = 68 ms.  Note we excluded the sa protocol of the Prisma scanner, due to severe mis-alignments.  For sa protocol, multiband-acquisition and stronger gradients shortened TE and improved both spatial and angular resolution per unit time. Additional b=0 s/mm\textsuperscript{2} images were acquired with TE and/or TR matching between protocols, as well as structural MPRAGEs for each scanner.  The data were corrected for EPI and eddy-current distortions, subject motion and gradient non-linearity, as well as co-registered together as in \cite{CSCPDMRIDH}. In addition, brain masks excluding the skull and background were provided for each subject and each acquisition.  The dataset size is representative of typical TH data-sets, where it is difficult to transport a large number of subjects from different centres and obtain multiple scans.  See figure-\ref{CDM} for a visualisation.

\noindent \\ \textbf{Processing and Training} Raw data were pre-processed to extract signal features by SH deconvolution -- variants of this technique are used as input-output common to the challenge and our comparison networks \cite{DIQT,HDMRID,SHRN}.  We employed  \cite{DEFODF,DIPY} where the 28 coefficients of the 6th order real-and-symmetric SH deconvolution were estimated from the normalized raw signal considered as separate channels of our data.  We first normalized the data, on each acquisition, per channel, to be mean $ 0 $ and standard deviation $ 1 $. 
\\ \indent Neural-network data-harmonisation transformations work patch-by-patch on an input image. Each voxel within the brain mask of the input acquisition defines a patch with that voxel at the centre. The corresponding target patch comes from the corresponding location in the target image.  For this paper we utilised input patch size as $ 11^{3} $, with target patch size $ 11^{3},19^{3} $, depending on the target resolution.  We then separated the patches into 90\%, 10\% training, test set.
\\ \indent Our implementation used PyTorch with Python, with a minibatch size of 12, ADAM optimizer, learning rate starting at $ 1E-4 $ and decaying by $ \sqrt{2} $ per $ 15 $ or $ 25 $ epochs, for a minimum of $ 50 $ total epochs \cite{CODE}.
\\ 
\\ \textbf{Comparison} We evaluate our approach via the tasks in MUSHAC \cite{CSCPDMRIDH}: to predict images from more modern platforms prisma\_st,  connectom\_st, connectom\_sa, from an older platform ge\_st (see figure-\ref{CDM}).  Our evaluation was performed on a local scale, as in \cite{CSCPDMRIDH}, we calculate the mean-squared-error (MSE) on the SH space.  This metric is not based on the raw dMRI and encompasses all key aspects of the diffusion signal (anistropy, mean diffusivity, principle orientation).
\\ \indent We select three baselines, representing state-of-the-art approaches \cite{CSCPHMSDMRIOCEV,MSDMRIEC}, that predict each acquisition using separate neural networks: CNN-RISH \cite{HDMRID}, DIQT \cite{DIQT}, SHResNet \cite{SHRN}.  We also consider simple MTL sub-networks from CPM \cite{CPM}, HNED \cite{HNED}, illustrated in the supplementary material, which were formulated for different tasks and inspired the MSP.  The super-resolution networks replaced a standard 3D convolution with a deconvolution / strided convolution layer, the connection networks are a stack of two 3D convolutions.  We then constructed the MSP by combining the best single-network approaches (bold in table-\ref{DIQT_SOA}).

\begin{figure}
\begin{center}
\includegraphics[width=\linewidth]{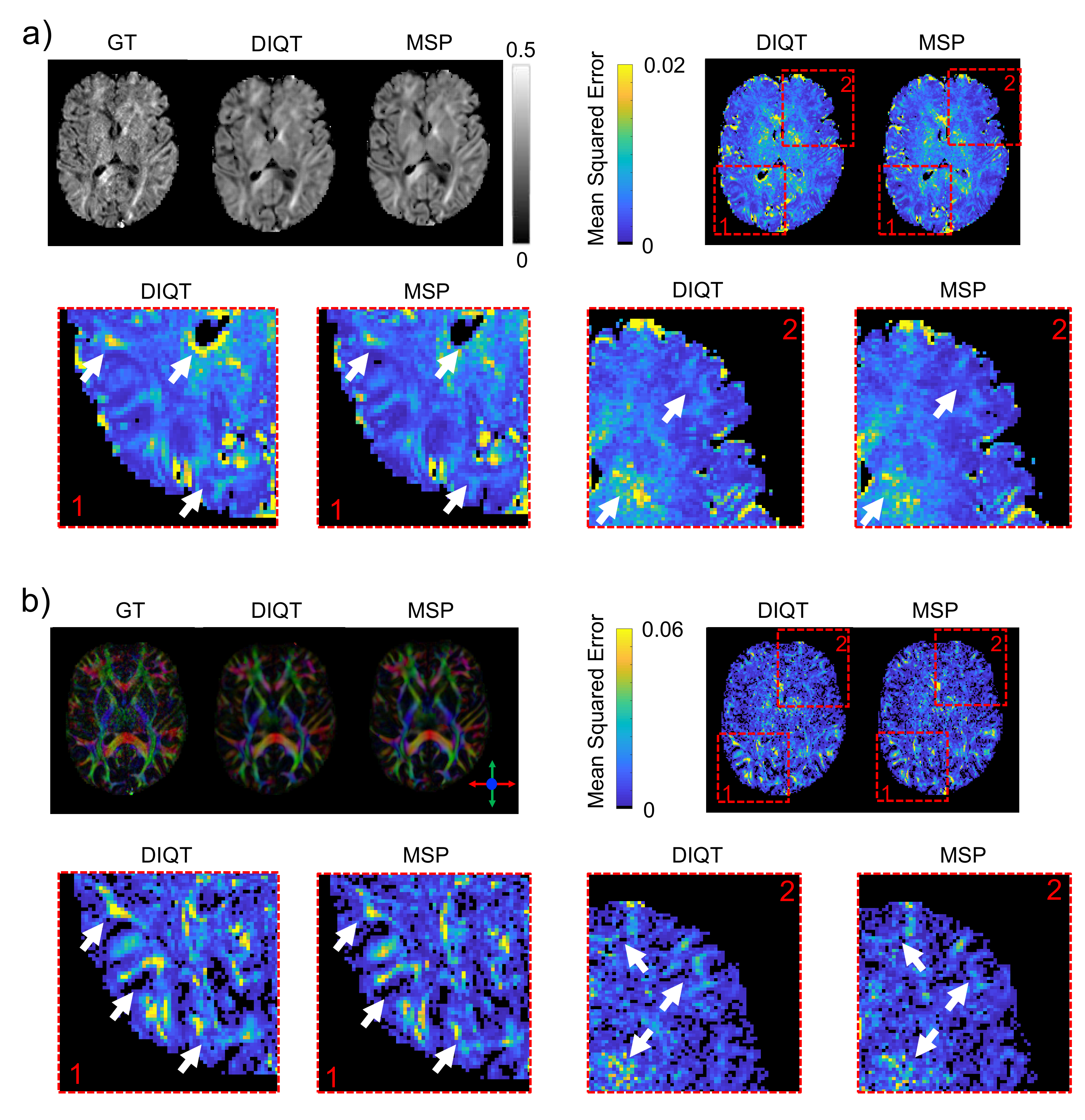}\caption{A qualitative comparison of the DIQT -- a single-network prediction, with our MSP network, compared to the Ground Truth (GT).  a) Comparison with the GT. The maps show the average of the first 6 SH coefficients. Quantitative maps of the MSE are also displayed in the second row. b) Comparison with the reference GT. The maps show the colour-coded fractional anisotropy (FA) from diffusion tensor imaging computed using FSL \cite{FSL}. Quantitative maps of the MSE are also displayed in the second row. }\label{Comparison_pic}
\end{center}
\end{figure}

\begin{table}[t]
\begin{center}
\footnotesize
\begin{tabular}{ |c|c||c|c|c |c| } 
 \hline
    Model    & Network Type      & prisma\_st      & connectom\_st                &  connectom\_sa        \\ 
\hline \hline     
   DIQT \cite{DIQT} & Single Network         &  76  ($ \pm $12)  &   59 ($ \pm 14 $)           &       \textbf{463} ($ \pm $ 123) \\  
  CNN-RISH \cite{HDMRID}   & Single Network   & 78 ($ \pm $11)    &   63 ($ \pm 7 $)            &       532 ($ \pm $ 144)  \\ %
  SHResNet \cite{SHRN}  & Single Network  & \textbf{70} ($ \pm $11)    &   \textbf{54} ($ \pm 7 $)   &       524 ($ \pm $ 112)   \\ %
\hline \hline
   CPM \cite{CPM}  & MTL   &   67  ($ \pm 12 $) &  48 ($ \pm 16 $)  &  417 ($ \pm 102 $)        \\   
   HNED \cite{HNED} & MTL   &   71 ($ \pm 12 $)  &  52 ($ \pm 14 $)  &  406 ($ \pm 98 $)         \\   
   MSP (Ours)  & MTL   &   \textbf{59} ($ \pm 10 $)  &  \textbf{43}  ($ \pm 9 $)  &  \textbf{374} ($ \pm 112 $)  \\  
   \hline
\end{tabular}
\end{center}\caption{Mean (std), of $ 278 $ patch MSEs.  We compare state-of-the-art single-network approaches, with MTL approaches, for three target platforms.  
}\label{DIQT_SOA}\label{MSP-2_table}\label{MSP-3_table}
\end{table}
\indent Table-\ref{DIQT_SOA} shows that MTL in general improves with respect to the single-network approaches, for all three target platforms. We use the Wilcoxon signed-rank test, a non-parametric statistical test, confirming that for all predicted platforms 
the differences in error scores are statistically significant, as $ p $ values obtained are all less than $ 10^{-5} $, after Bonferroni correction.
\\ \indent Figure-\ref{Comparison_pic} shows qualitative results revealing differences in results from DIQT and MSP for connectom\_sa, on a subject excluded from the training set.  Improvements from MSP over  DIQT are subtle but visible on both the SH and direction-encoded colour maps, e.g. MSP reduces errors directly underneath the left ventricle.  Quantitatively in that figure, MSP improves the global SH MSE by 5\% and the global direction-encoded colour MSE by 11\%.

\section{Conclusion and Future Work}\label{DAFW}

In this paper we demonstrated how MTL approaches such as the MSP, increase the predictive power of deep learning in DH and improve over state-of-the-art single network approaches.
\\ \indent There are many potential directions of future work.  Here we test just one MSP architecture, but many variations of deeper structures with more sophisticated connections between tasks are possible.  For example, the MSP chooses $ z_{i} $ to be the
the deepest feature of $ \mathcal{N}_{iT} $, in future work we could experiment with connecting different layers.
\\ \indent One key problem in evaluating DH performance is residual misalignment of input and output images. This can mask strong performance when using simple global comparison metrics.  It may be useful to devise alternative evaluation metrics that are robust to misalignment, or by moving away from direct image evaluation and analyzing downstream processes.
\\ \indent In this paper, we demonstrated the MSP via a diffusion MRI data set.  However the same architecture extends naturally to harmonizing data between multiple different modalities, within or beyond MRI.

\section*{Acknowledgements}
We thank: Tristan Clark and the HPC team (James O'Connor, Edward Martin) and the organizers of the 2017 MUSHAC challenge (Francesco Grussu, Enrico Kaden,  Lipeng Ning, Jelle Veraart).  This work was supported by an EPRSC and Microsoft scholarship and EPSRC grants M020533 R006032 R014019 and the NIHR UCLH Biomedical Research Centre.  The data were acquired at the UK National Facility for In Vivo MR Imaging of Human Tissue Microstructure located in CUBRIC funded by the EPSRC (grant EP/M029778/1), and The Wolfson Foundation. Acquisition and processing of the data was supported by a Rubicon grant from the NWO (680-50-1527), a Wellcome Trust Investigator Award (096646/Z/11/Z), and a Wellcome Trust Strategic Award (104943/Z/14/Z).

\bibliographystyle{splncs}

\section*{Supplementary Material}

\begin{figure}[H]
\begin{center}
\includegraphics[width=\textwidth]{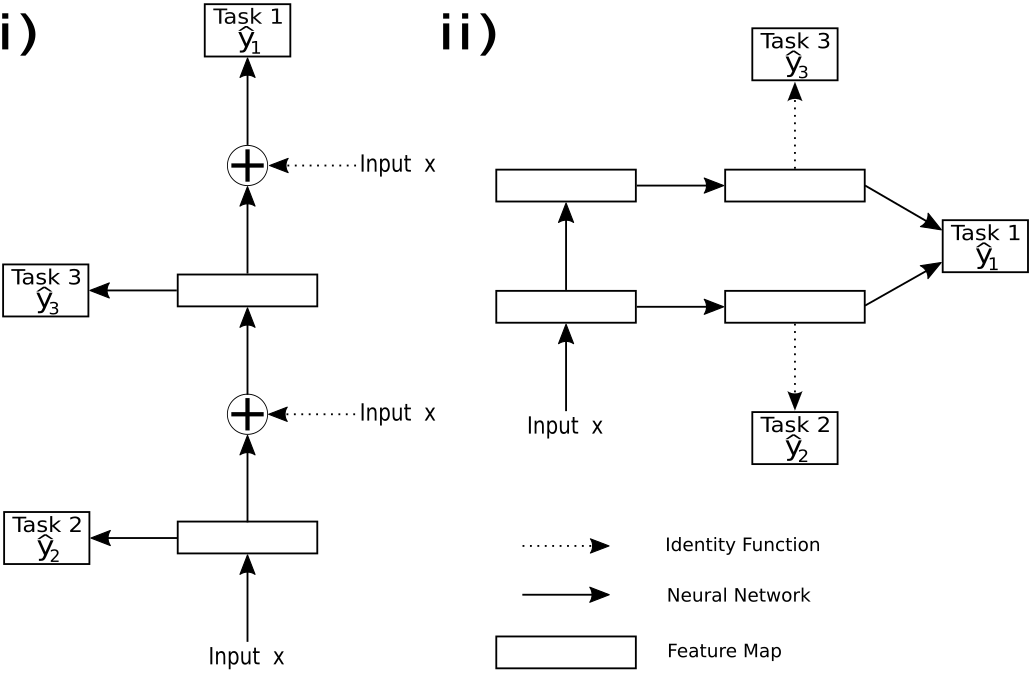}\caption{An illustration of two MTL approaches that inspired the MSP, with input, target platform 0,1 and other platforms 2,3.  The input patch is $ x $, the prediction patch of platform $ j $ is $ \widehat{y}_{j} $. i) Denoted as CPM is from \cite{CPM} ii) Denoted as HNED is from \cite{HNED}. }\label{Supplementary_fig}
\end{center}
\end{figure}

\end{document}